\documentclass[runningheads]{llncs}
\usepackage[T1]{fontenc}
\usepackage{xcolor}
\usepackage{multirow}
\usepackage{graphicx}
\usepackage{amsmath}
\usepackage{booktabs}
\usepackage{color}
\usepackage[colorlinks=true,linkcolor=blue,urlcolor=blue,citecolor=blue]{hyperref}

\begin{document}
\title{CAF-I: A Collaborative Multi-Agent Framework for Enhanced Irony Detection with Large Language Models}
\titlerunning{CAF-I: Collaborative Multi-Agent Irony Detection Framework}

\author{
  Ziqi Liu\inst{1} and Ziyang Zhou\inst{1} \thanks{\textbf{Z. Liu and Z. Zhou contributed equally to this work.}}, Mingxuan Hu\inst{1}}
\authorrunning{Z. Liu and Z. Zhou, M. Hu}

\institute{
  Xi'an Jiaotong-Liverpool University\\
  \email{\{Ziqi.Liu22,ZiYang.Zhou22,Mingxuan.Hu22\}@student.xjtlu.edu.cn}
}
\maketitle           
\begin{abstract}
Large language model (LLM) have become mainstream methods in the field of sarcasm detection. However, existing LLM methods face challenges in irony detection, including: \textbf{1. single-perspective limitations}, \textbf{2. insufficient comprehensive understanding}, and \textbf{3. lack of interpretability}. This paper introduces the Collaborative Agent Framework for Irony (\textbf{CAF-I}), an LLM-driven multi-agent system designed to overcome these issues. CAF-I employs specialized agents for Context, Semantics, and Rhetoric, which perform multidimensional analysis and engage in interactive collaborative optimization. A Decision Agent then consolidates these perspectives, with a Refinement Evaluator Agent providing conditional feedback for optimization. Experiments on benchmark datasets establish CAF-I's state-of-the-art zero-shot performance. Achieving SOTA on the vast majority of metrics, CAF-I reaches an average \textbf{Macro-F1 of 76.31\%}, a \textbf{4.98\%} absolute improvement over the strongest prior baseline. This success is attained by its effective simulation of human-like multi-perspective analysis, enhancing detection accuracy and interpretability.
\end{abstract}

\section{Introduction}
The rapid development of social media has fostered increasingly diverse and semantically complex forms of textual expression \cite{kader2022computational} highlights the challenge of irony detection. Irony, a rhetorical device contrasting stated and intended meaning \cite{booth1974rhetoric},  is crucial for NLP systems as its presence alters textual interpretation, impacting applications like sentiment analysis and content moderation \cite{reyes2013multidimensional}. However, detection is complicated by context dependency, obscure authorial intent, and varied rhetorical strategies \cite{wallace2015sparse}.

Initial efforts used traditional machine learning with handcrafted features \cite{reyes2013multidimensional}, often failing to capture irony's subtleties. Deep learning methods improved semantic representation but struggled with the implicit and contextual nature of ironic utterances \cite{ghosh2016fracking,baziotis2018ntua}. Consequently, the emergence of LLM has presented novel avenues, exhibiting substantial potential in irony detection, typically via fine-tuning or advanced prompt engineering techniques \cite{yao2024sarcasm}.

\begin{figure}
    \vspace{-.4cm}
    \centering
    \includegraphics[width=0.8\linewidth]{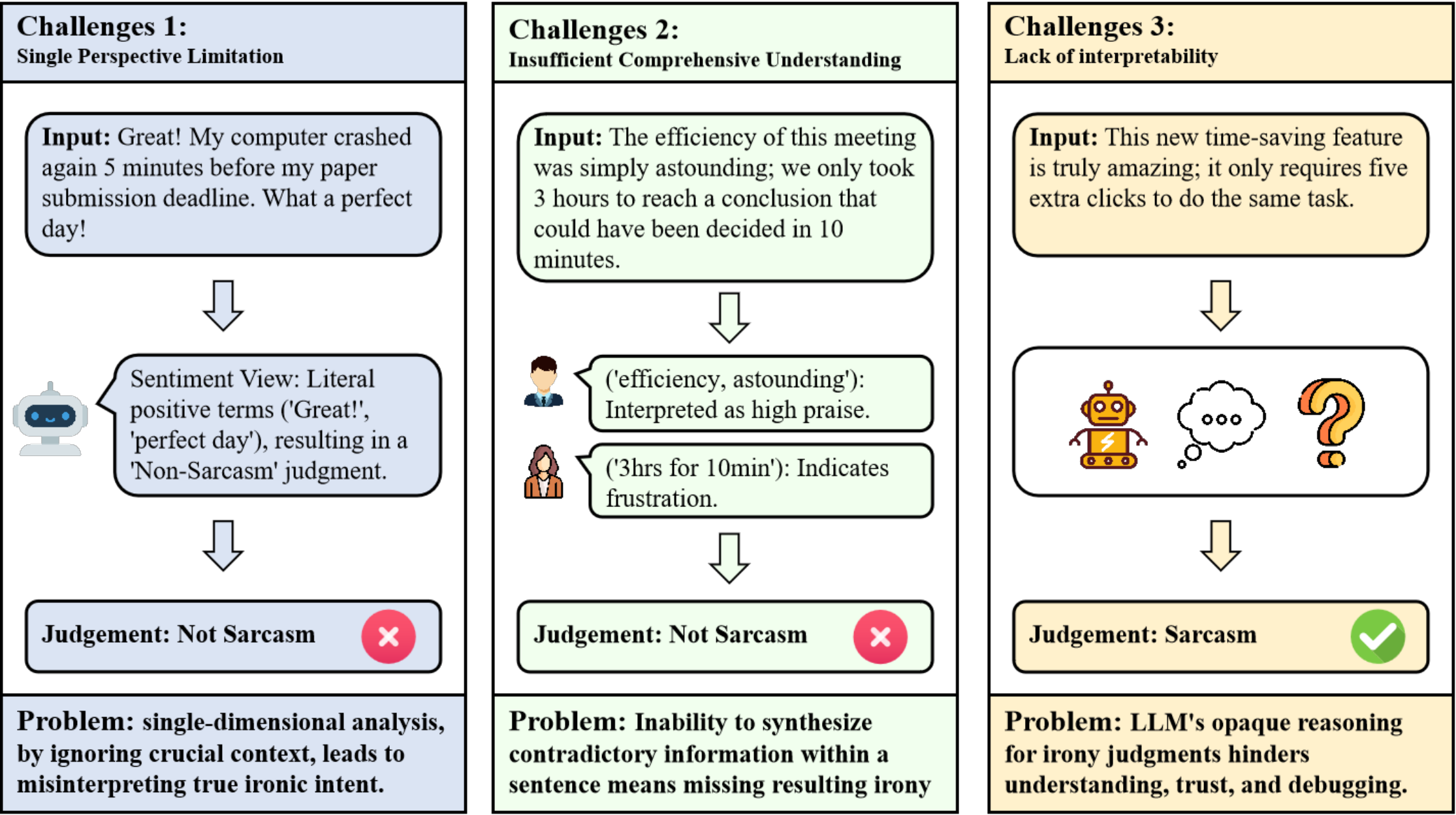}
    \caption{Examples illustrating three primary challenges for LLM in sarcasm detection: single perspective limitation, insufficient comprehensive understanding, and lack of interpretability.}
    \label{fig:intro}
    \vspace{-.5cm}
\end{figure}

Despite this significant promise and their advanced capabilities, LLM encounters a distinct set of challenges when specifically applied to the nuanced task of irony detection, as illustrated in Figure \ref{fig:intro}. Key among these challenges are:
\begin{enumerate}
    \item \textbf{Single-Perspective Limitation:} As single-model predictors, LLM have limited capacity for multidimensional collaborative reasoning, struggling to collaboratively synthesize diverse analytical insights required for deconstructing complex irony.
    \item \textbf{Insufficient Comprehensive Understanding:} LLM struggle to holistically integrate diverse informational cues, such as context, semantics, and rhetoric, to achieve human-like deep comprehension of ironic intent.
    \item \textbf{Lack of Interpretability:} Opaque reasoning process in irony judgments hinders the understanding, trust, and debugging of decision-making mechanisms.
\end{enumerate}

Motivated by Chain-of-Thought (CoT) prompting \cite{wei2022chain} and multi-agent cooperation principles \cite{liu2023agentbench}, this paper introduces CAF-I, a novel LLM-driven multi-agent framework to address existing limitations in irony detection by simulating human-like multidimensional analytical processes. CAF-I initially performs multi-perspective parallel analysis from three core angles: contextual consistency, deep semantic logic, and rhetorical function. Subsequently, an interactive collaborative optimization mechanism facilitates information sharing and debate among these perspectives. Finally, an integrated decision mechanism allows a central decision-making agent to consolidate these viewpoints and apply conditional optimization triggered by evaluative feedback, forming the paper's primary contributions:
\begin{itemize}
    \item Introduces CAF-I, a novel LLM-driven multi-agent framework significantly improving irony detection through specialized agent collaboration and phased optimization.
    \item Achieves state-of-the-art zero-shot performance across multiple benchmark datasets.
    \item Its unique architecture offers clear decision pathways and error correction capabilities, enhancing interpretability.
    \item Extensive experiments validate its component necessity, overall robustness, and innovation, establishing CAF-I as a pioneering application of specialized LLM agents in structured collaboration for irony detection.
\end{itemize}

\section{RELATED WORK}
\subsection{Irony Detection}
Irony detection in NLP remains challenging due to its inherent complexity, context dependence, and contradictory nature. Early machine learning relied on hand-crafted features, including lexical cues \cite{davidov2010semi} and sentiment lexicons \cite{reyes2013multidimensional}, but inadequately captured irony's subtleties. Deep learning techniques offered significant improvements: Word embeddings such as Word2Vec \cite{mikolov2013efficient} and GloVe \cite{pennington2014glove} enabled richer semantic representations. Subsequently, Convolutional Neural Networks \cite{poria2016deeper}, recurrent neural networks, and particularly LSTM architectures \cite{zhang2016tweet} learned hierarchical and sequential features, while Graph Neural Networks \cite{liang2022multi} captured structural dependencies to improve performance. Nonetheless, these deep learning approaches still faced difficulties with irony's implicit meanings and complex rhetoric, motivating the exploration of more advanced models.

Recent advances in large language models have introduced prompt-based and zero-shot learning methods, allowing models like GPT-4o to interpret irony with minimal reliance on labeled data \cite{yao2024sarcasm}. By leveraging task-specific prompts, these models have demonstrated enhanced context-aware irony classification. However, existing LLM-based approaches still operate as single-model predictors, lacking the capacity for multi-perspective reasoning.

\subsection{CoT Prompt Engineering}
To address the limitations of direct predictions in complex reasoning tasks, CoT prompting emerged as a structured technique that enhances language models by guiding them through intermediate reasoning steps \cite{wei2022chain}. Despite its effectiveness, CoT initially depended on manually created prompts, which constrained its adaptability. Subsequent innovations, such as Auto-CoT, addressed this by automating the creation of reasoning chains \cite{zhang2022automatic}. Further extensions, such as Tree-of-Thought (ToT), expanded reasoning by enabling the exploration of multiple pathways \cite{yao2024sarcasm}, while Graph-of-Thought (GoT) introduced a structured approach where reasoning steps are represented as interconnected nodes in a graph \cite{besta2024graph}. These advancements directly influenced our multi-agent framework, where specialized agents leverage structured reasoning to collaboratively enhance irony detection.

\subsection{Multi-agent Cooperation}
Multi-agent frameworks effectively leverage collaborative interactions among specialized agents for complex tasks, with studies exploring forms like deliberation, structured debates, and dialogues. Debating frameworks, for instance, improve factual accuracy and solution diversity in complex reasoning \cite{du2023improving}. Others, such as CAMEL \cite{li2023camel}, use role-play to simulate nuanced human behaviors; MathChat \cite{wu2023mathchat} employs structured dialogues for intricate tasks; and AutoGen \cite{wu2023autogen} shows multi-agent versatility through customizable structures. Inspired by these methodologies, our framework uses specialized LLM-based agents for contextual, semantic, and rhetorical analysis via CoT reasoning. A final decision agent integrates their outputs, enhancing irony detection's robustness, accuracy, and interpretability.

\section{Methodology}
This section details our methodology for irony detection, centered on a novel multi-agent collaborative framework driven by LLM, termed the CAF-I. This framework comprises a Context Agent, a Semantic Agent, a Rhetoric Agent, a Decision Agent, and a Refinement Evaluator Agent.

CAF-I aims to address single-model limitations in irony detection via multi-agent collaboration. As irony comprehension demands nuanced understanding of context, semantics, and rhetoric, CAF-I employs dedicated agents for these dimensions. A Decision Agent aggregates these analyses, refined by the conditional feedback of the refinement evaluator agent, to achieve robust and accurate detection.

\subsection{Problem Definition}
Given an input text collection $X = \{x_i\}_{i=1}^n$, the primary objective is to accurately identify ironic expressions within each text $x_i$. This task is formalized as a binary classification problem. For each input text $x_i \in X$, the goal is to predict a Boolean label $y_i \in \{\text{Ironic}, \text{Non-Ironic}\}$, indicating the presence or absence of irony. This work aims to develop an advanced LLM-powered system to address this challenge.

\subsection{Overall Workflow}
\begin{figure*}
    \centering
    \includegraphics[width=1.0\linewidth]{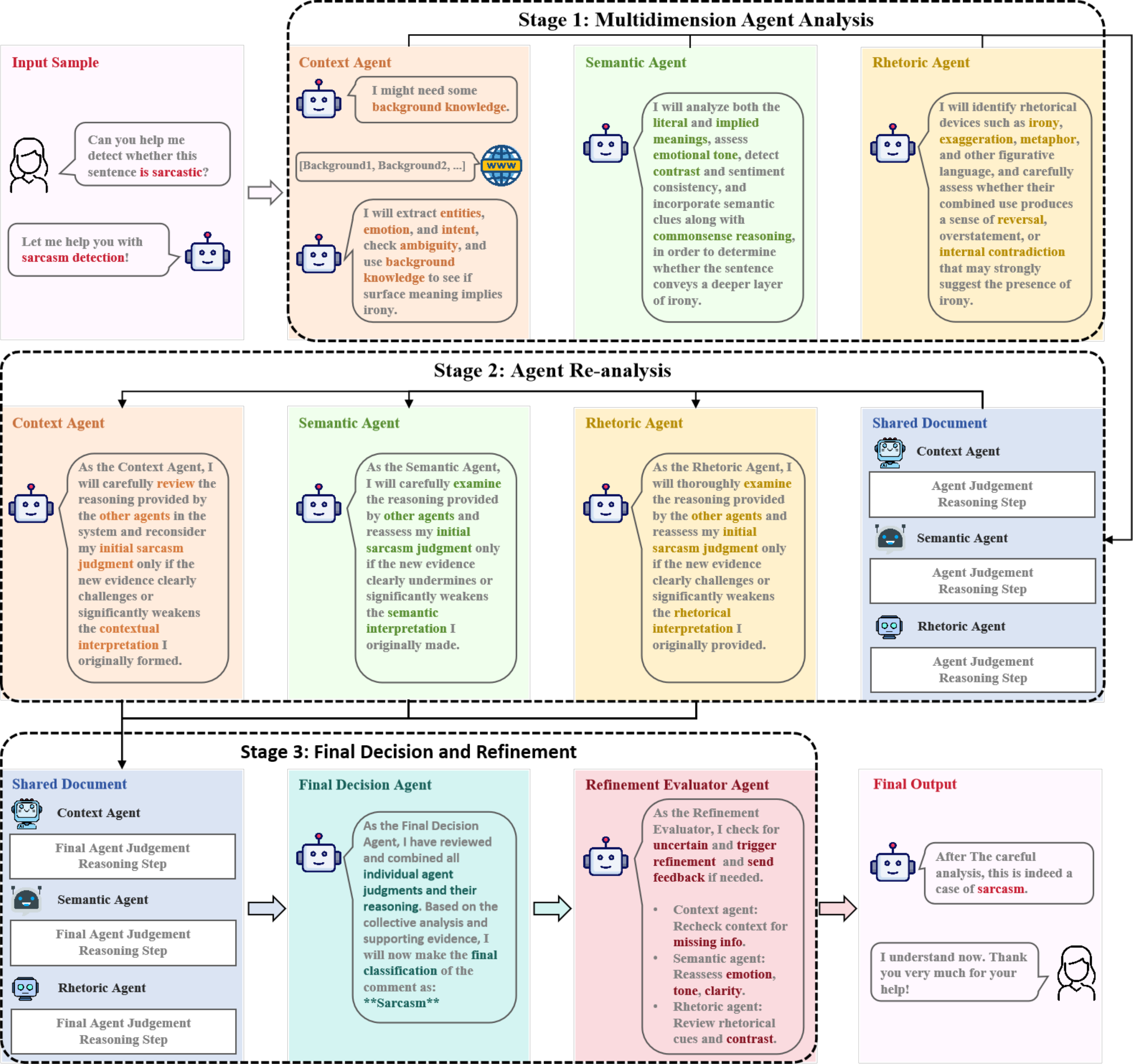}
    \caption{The Overall architecture of CAF-I. The framework includes three specialized agents, Context, Semantic, and Rhetoric, which conduct multi-round analysis and optimization of the input text. The Decision Agent integrates the preliminary classification, and the Evaluator Agent determines if a round of feedback optimization is needed.}
    \label{fig:structure}
    \vspace{-.6cm}
\end{figure*}

The CAF-I inference workflow, depicted in Figure \ref{fig:structure}, unfolds in key stages. Initially, three specialized analysis agents—Context (CA), Semantic (SA), and Rhetoric (RA)—independently provide first-round assessments and reasoning. These outputs are shared for a collaborative reanalysis, where agents refine their judgments considering peer insights. A Decision Agent (DA) then aggregates these refined multiperspective judgments into an initial irony classification. Finally, a Refinement Evaluator Agent (REAgent) reviews this output and, if needed, triggers a single conditional refinement loop with targeted feedback to the analysis agents, aiming to improve the final classification within at most one iteration. \textbf{The detailed design of each agent mentioned above is presented in \ref{agent design}.}

\subsection{Agent Design}
\label{agent design}
\subsubsection{Context Agent}
\paragraph{(1) Role}
The Context Agent identifies irony via semantic analysis of input $x_i$ and strategic external knowledge use, detecting inconsistencies between literal and contextual meanings.

\paragraph{(2) Workflow}
Upon receiving an input text $x_i$, the Context Agent extracts internal contextual features using an LLM. This involves identifying key named entities $E = \{e_1, ..., e_k\}$, their semantic relationships $R = \{r_1, ..., r_l\}$, and the overarching theme $T$. These components collectively define the internal context, denoted as $\mathcal{C}_{int} = (E, R, T)$. The overall extraction process can be formulated as:
\begin{equation}
\mathcal{C}_{int} = \text{ExtractInternalContext}{\text{LLM}}(x_i)
\end{equation}
Subsequently, the CA, guided by an LLM, assesses the need for external knowledge based on the ambiguity of $x_i$ and the novelty of $\mathcal{C}_{int}$. If deemed beneficial, a search query $q$ is formulated using $q = g_{\text{LLM}}(\text{keywords}(x_i))$. Retrieval via an external search API yields a set of documents $D = \{d_1, ..., d_p\}$. These are summarized into an external context representation $S$. 

Finally, a specialized LLM-based classifier, $\text{IronyClassifier}_{\text{LLM}}$, integrates the original input text $x_i$, the extracted internal context $\mathcal{C}_{int}$, and the summarized external knowledge $S$ (if retrieved) to directly output a binary decision $y_1 \in \{\text{Ironic}, \text{Non-Ironic}\}$ regarding the presence of irony, along with associated reasoning. This process can be represented as:
\begin{equation}
(y_1, R_1) = \text{IronyClassifier}_{\text{LLM}}(x_i, E, R, T, S)
\end{equation}

\subsubsection{Semantic Agent}
\paragraph{(1) Role}
The Semantic Agent employs a CoT process for detailed semantic analysis to identify irony. It focuses on the interplay of emotional tone, implied meaning, and commonsense expectations to detect nuanced incongruities.

\paragraph{(2) Workflow}
The workflow initiates with an LLM performing semantic parsing of the input $x_i$ to discern its literal meaning $M_{lit}(x_i)$ from pragmatically implied intent $I_{imp}(x_i)$:
\begin{equation}
    (M_{lit}(x_i), I_{imp}(x_i)) = \text{SemanticParse}_{\text{LLM}}(x_i)
\end{equation}
Subsequently, as part of its CoT reasoning, the agent identifies the expressed emotion $E_{exp}(x_i)$:
\begin{equation}
    E_{exp}(x_i) = \text{EmotionClassifier}_{\text{LLM}}(x_i)
\end{equation}
It then evaluates the consistency between $E_{exp}(x_i)$ and the contextually expected emotion $E_{ctx}(x_i)$. Concurrently, the SA assesses the text's consistency with commonsense knowledge $\mathcal{K}_{cs}$:
\begin{equation}
\begin{aligned}
\text{Consistency}(x_i, \mathcal{K}_{cs}) &= \text{Verifier}_{LLM}(x_i, \mathcal{K}_{cs}) \\
&\in \{\text{Consistent}, \text{Inconsistent}\}.
\end{aligned}
\end{equation}
The SA's final irony classification $y_2$ and its supporting explanation $R_2$ are generated by an LLM that holistically reasons over the outcomes of semantic parsing, the evaluation of emotional consistency, and commonsense reasoning through a CoT process:
\begin{equation}
\begin{aligned}
(y_2, R_2) &= \text{SemanticIronyDetector}_{\text{LLM}}(M_{lit}(x_i), I_{imp}(x_i) \\
&\qquad E_{exp}(x_i), E_{ctx}(x_i), \text{Consistency}(x_i, \mathcal{K}_{cs}))
\end{aligned}
\end{equation}

\subsubsection{Rhetoric Agent}
\paragraph{(1) Role}
The Rhetoric Agent identifies irony by analyzing rhetorical devices within the input text $x_i$. It evaluates their interplay with communicative intent, using CoT prompting, to assess irony based on linguistic form.

\paragraph{(2) Workflow}
Employing an LLM with CoT, the RA first identifies rhetorical devices $D_r$ present in the text:
\begin{equation}
    D_r = \text{RhetoricalDeviceDetector}_{\text{LLM}}(x_i) = \{d_{r1}, d_{r2}, ..., d_{rm}\}
\end{equation}
For each detected device $d_{rj} \in D_r$, the LLM explains its rhetorical function $E_{rj}$:
\begin{equation}
    E_{rj} = \text{ExplainRhetoricalFunction}_{\text{LLM}}(d_{rj})
\end{equation}
Subsequently, the RA then assesses if the overall rhetorical structure, guided by $D_r$ and their explanations $\{E_{rj}\}_{j=1}^{m}$, indicates irony, potentially evidenced by contradictions, reversals, or exaggerations. An LLM-based classifier generates the final judgment $y_3$ and explanation $R_3$ from these inputs:
\begin{equation}
    (y_3, R_3) = \text{RhetoricalIronyClassifier}_{\text{LLM}}(D_r, \{E_{rj}\}_{j=1}^{m})
\end{equation}

\subsubsection{Decision Agent}
\paragraph{(1) Role}
The final Decision Agent produces a robust and interpretable irony classification $y$ by consolidating the analysis agents' second-round outputs, comprising their judgments and reasoning traces.

\paragraph{(2) Workflow}
The Decision Agent initiates its process by collecting the second-round outputs from the three primary analysis agents: their binary judgments $y_1^{(2)}, y_2^{(2)}, y_3^{(2)}$ and their corresponding reasoning traces $R_1^{(2)}, R_2^{(2)}, R_3^{(2)}$. 

The DA employs a hierarchical strategy. If agents achieve consensus, their unanimous judgment, denoted as $y^*$, is adopted:
\begin{equation}
    y = y^*, \quad \text{if } y_1^{(2)} = y_2^{(2)} = y_3^{(2)}
\end{equation}
If a two-agent majority exists, the decision is made by majority vote:
\begin{equation}
\begin{aligned}
y &= \text{majority}(y_1^{(2)}, y_2^{(2)}, y_3^{(2)}), \quad \text{if } (y_1^{(2)} = y_2^{(2)} \neq y_3^{(2)}) \\
&\qquad \lor (y_1^{(2)} = y_3^{(2)} \neq y_2^{(2)}) \lor (y_2^{(2)} = y_3^{(2)} \neq y_1^{(2)})
\end{aligned}
\end{equation}
In cases of complete disagreement, an LLM analyzes the agents' reasoning traces ($R_i^{(2)}$) for clarity, coherence, and relevance, adopting the judgment supported by the most compelling argument.

The DA ultimately outputs the final binary classification $y \in \{\text{Ironic}, \text{Non-Ironic}\}$ with a synthesized justification derived from the most persuasive reasoning.

\subsubsection{Refinement Evaluator Agent}

\paragraph{(1) Role.}
The Refinement Evaluator Agent operates subsequently, once per inference, to assess the preliminary decision $s$ and explanation $\mathcal{E}$ from the Decision Agent. It determines if a single refinement iteration is necessary ($R_{needed}$) and generates hypothetical feedback as a triplet $(f_{CA}, f_{SA}, f_{RA})$ for upstream agents to improve reliability.

\paragraph{(2) Workflow.}
The REAgent first evaluates $s$ and $\mathcal{E}$ for internal quality, inferring a confidence level $C_{RE} \in \{\text{High, Medium, Low}\}$, and checks for strong contradictions, yielding $F_{contra} \in \{\text{Yes, No}\}$. The necessity of refinement, $R_{needed}$, is determined by:
\begin{equation} \label{eq:re_needs_reflection_final}
R_{needed} =
\begin{cases}
  \text{"true"} & \text{if } C_{RE} \in \{\text{Low}\} \lor F_{contra} = \text{Yes} \\
  \text{"false"} & \text{otherwise}
\end{cases}
\end{equation}
If $R_{needed} = \text{"true"}$, the REAgent generates concise textual feedback components $f_{CA}, f_{SA}, f_{RA}$, suggesting attention points based on weaknesses inferred from $s$ and $\mathcal{E}$. The REAgent outputs $R_{needed}$ and the feedback triplet, signaling the control logic to either accept the initial result or perform a single, final refinement iteration.

\section{EXPERIMENT}
In this section, we try to answer the following research questions:
\begin{itemize}
    \item \textbf{RQ1:} How does CAF-I perform on benchmark datasets against diverse existing irony detection methods?
    \item \textbf{RQ2:} What are the performance contributions of CAF-I's core components?
    \item \textbf{RQ3:} How robust is CAF-I's architecture with different LLM backbones, and how does it compare to standard prompting of these LLM?
    \item \textbf{RQ4:} How interpretable is CAF-I, and how valuable are its agents' intermediate reasoning steps for detection performance?
    \item \textbf{RQ5:}  How does CAF-I's inference efficiency compare to other competitive CoT-based LLM approaches?
\end{itemize}

\subsection{Experiment Setup}

\subsubsection{Datasets}
We evaluate CAF-I's efficacy and generalizability using four established sarcasm detection benchmarks:
\begin{itemize}
    \item \textbf{IAC-V1} \cite{lukin2013really}: This dataset comprises comments from online political debates, offering examples of sarcasm embedded within argumentative discourse.
    \item \textbf{IAC-V2} \cite{oraby2017creating}: An extension of IAC-V1, providing a larger, more diverse collection of sarcastic/non-sarcastic statements from similar online discussion forums.
    \item \textbf{MuSTARD} \cite{castro2019towards}: Sourced from popular television shows, MuSTARD provides conversational context for each utterance. 
    \item \textbf{SemEval-2018 Task 3} \cite{van2018semeval}: Consists of English tweets annotated for irony, representing challenges of social media text like brevity, informal language, and implicit contextual cues for ironic intent.
\end{itemize}

Table~\ref{table:dataset_stats} summarizes detailed dataset statistics. Their diversity in domains and challenges enables robust framework evaluation.

\begin{table}[!htp]
\vspace{-.2cm}
    \centering
    \caption{Overview of the benchmark datasets used for evaluating irony detection.}
    \label{table:dataset_stats}
    \renewcommand{\arraystretch}{0.8} 
    \tiny
    \resizebox{0.7\textwidth}{!}{%
    \begin{tabular}{lccccc}
        \toprule
        Dataset & Year & Size & Avg. Length & Domain & Context \\
        \midrule
        IAC-V1 & 2013 & 320 & 68 & Debate & No \\
        IAC-V2 & 2016 & 1042 & 43 & Debate & No \\
        MuSTARD & 2019 & 784 & 14 & Dialogue & Yes \\
        SemEval-2018 & 2018 & 183 & 14 & Twitter & No \\
        \bottomrule
    \end{tabular}
    }
    \vspace{-.6cm}
\end{table}
\subsubsection{Evaluation Metrics}

We evaluate model performance using \textbf{Accuracy} for overall correctness and the \textbf{Macro-F1 score} as the primary metric, following standard practices in the field \cite{yao2024sarcasm}. Macro-F1 averages the F1-scores for ironic and non-ironic classes independently, providing a balanced measure suitable for potentially imbalanced datasets common in irony detection.

\subsubsection{Experiment Details}

The \textbf{GPT-4o} model serves as the LLM backbone for all agents within the proposed CAF-I framework, accessed via the official OpenAI API. Its selection ensures state-of-the-art baseline capabilities in natural language understanding and reasoning. To enhance reproducibility, a temperature setting of 0 was used for all API interactions.

\subsubsection{Comparison Baselines}
To contextualize CAF-I's performance, we compare it against a comprehensive suite of baselines across three categories:
\textbf{LLM-based:} Approaches from the SarcasmCue framework \cite{yao2024sarcasm}, including GPT-4o Zero-Shot and three advanced prompting strategies: Chain of Contradiction, Graph of Cues, and Bagging of Cues. Performance figures are cited from their reported results.
\textbf{Fine-tuned PLMs:} Standard Pre-trained Language Models such as BERT-base-uncased \cite{devlin2019bert} and RoBERTa-base \cite{liu2019roberta}, fine-tuned on target datasets.
\textbf{Deep Learning Methods:} Influential deep learning methods with task-specific architectures, namely MIARN \cite{tay2018reasoning}, SAWS \cite{pan2020modeling}, and DC-Net \cite{liu2021dual}. Results are sourced from literature with aligned evaluation settings \cite{qiu2024detecting,zhang2024sarcasmbench,tay2018reasoning,xue2024breakthrough,pan2020modeling}.

\subsection{Overall Performance Comparison}

\begin{table*}[!htp]
\vspace{-.4cm}
    \centering
    \caption{Overall performance comparison across four benchmark datasets. All LLM strategies are zero-shot. Acc. denotes Accuracy and Ma-F1 signifies Macro-F1. Best results are presented in \textbf{bold}, second-best are \underline{underlined}. Scores are reported as \%.}
    \label{tab:main_results}
    \renewcommand{\arraystretch}{0.85} 
    \scriptsize
    \resizebox{\textwidth}{!}{
    \begin{tabular}{l|cc|cc|cc|cc|cc}
        \toprule
            \multirow{2}{*}{\textbf{Method}} & \multicolumn{2}{c}{\textbf{IAC-V1}} & \multicolumn{2}{c}{\textbf{IAC-V2}} & \multicolumn{2}{c}{\textbf{MuSTARD}} & \multicolumn{2}{c}{\textbf{SemEval-2018}} & \multicolumn{2}{c}{\textbf{Avg.}}\\
        \cmidrule(lr){2-3} \cmidrule(lr){4-5} \cmidrule(lr){6-7} \cmidrule(lr){8-9} \cmidrule(lr){10-11}
         & Acc. & Ma-F1 & Acc. & Ma-F1 & Acc. & Ma-F1 & Acc. & Ma-F1 & Acc. & Ma-F1\\
        \midrule
        MIARN \cite{tay2018reasoning} & 63.21 & 63.18 & 72.75 & 72.75 & 64.60 & 63.90 & 68.50 & 67.80 & 67.26 & 66.91 \\
        SAWS \cite{pan2020modeling} & 66.13 & 65.60 & 76.20 & 76.20 & 69.71 & 70.95 & 69.90 & 68.90 & 70.48 & 70.41\\
        DC-Net \cite{liu2021dual} & 66.50 & 66.40 & \textbf{78.00} & \textbf{77.90} & \underline{71.28} & \underline{71.43} & 70.80 & 69.60 & \underline{71.64} & \underline{71.33} \\
        BERT \cite{devlin2019bert} & 65.30 & 65.20 & 76.40 & 76.20 & 64.30 & 64.30 & 69.90 & 68.40 & 68.97 & 68.52 \\
        RoBERTa \cite{liu2019roberta} & 70.10 & 69.90 & 76.60 & 76.70 & 66.10 & 66.00 & 70.20 & 69.10 & 70.75 & 70.42\\
        GPT-4o \cite{yao2024sarcasm} & 70.63 & 70.05 & 73.03 & 71.99  & 67.24 & 65.79 & 64.03 & 63.17 & 68.73 & 67.75\\
        GPT-4o+CoC \cite{yao2024sarcasm} & \underline{72.19} & \underline{71.52} & 73.36 & 72.31 & 69.42 & 68.48 & 70.79 & 70.60 & 71.44& 70.73 \\
        GPT-4o+Goc\cite{yao2024sarcasm} & 65.00 & 62.91 & 64.97 & 61.30 & 70.69 & 69.91 & \underline{74.03} & \underline{74.02}  &  68.67& 67.04\\
        GPT-4o+Boc\cite{yao2024sarcasm} & 68.75 & 67.36 & 71.35 & 69.39 & 69.42 & 68.45 & 62.12 & 61.85 & 67.91 & 66.76\\
        \midrule
        \textbf{CAF-I (Ours)} & \textbf{73.75} & \textbf{73.71} & \underline{77.87} & \underline{76.82} & \textbf{75.21} & \textbf{74.73} & \textbf{80.73} & \textbf{79.99} & \textbf{76.89}& \textbf{76.31} \\
        \bottomrule
    \end{tabular}
    }
    \vspace{-.2cm}
\end{table*}
The overall performance comparison, with results detailed in table \ref{tab:main_results}, reveals several key insights. 

\textbf{Our Framework Achieves New SOTA.} Our proposed CAF-I framework demonstrates clear superiority, establishing a new state-of-the-art. It achieves the highest average Accuracy of 76.89 percent and an average Macro-F1 of 76.31 percent across all benchmarks, consistently outperforming other methods on most datasets and metrics. This underscores the efficacy of CAF-I's structured multi-agent collaborative reasoning for irony detection. 

\textbf{CoT in CAF-I Outperforms Simpler Prompting.} Within LLM-based approaches, our CAF-I framework clearly demonstrates the significant advantage of its CoT reasoning over simpler zero-shot prompting. Specifically, CAF-I's multi-agent CoT process achieves an average Macro-F1 of 76.31 percent, substantially outperforming the GPT-4o zero-shot baseline's average Macro-F1 of 67.75 percent. This highlights the substantial benefits derived from employing a sophisticated CoT methodology like that inherent in CAF-I. 

\textbf{LLM-based Methods Outperform Non-LLM Approaches.} Finally, LLM-based methods, on the whole, exhibit stronger performance compared to non-LLM approaches. Advanced LLM systems like CAF-I, with an average Macro-F1 of 76.31 percent, substantially outperform leading non-LLM methods; for example, the fine-tuned RoBERTa achieved an average Macro-F1 of 70.42 percent, and the traditional deep learning model DC-Net reached 71.33 percent. This indicates the enhanced capability of sophisticated LLM architectures for complex irony understanding over earlier paradigms.

\subsection{Ablation Study}
\begin{figure}[!htp]
    \centering
    \includegraphics[width=1.0\linewidth]{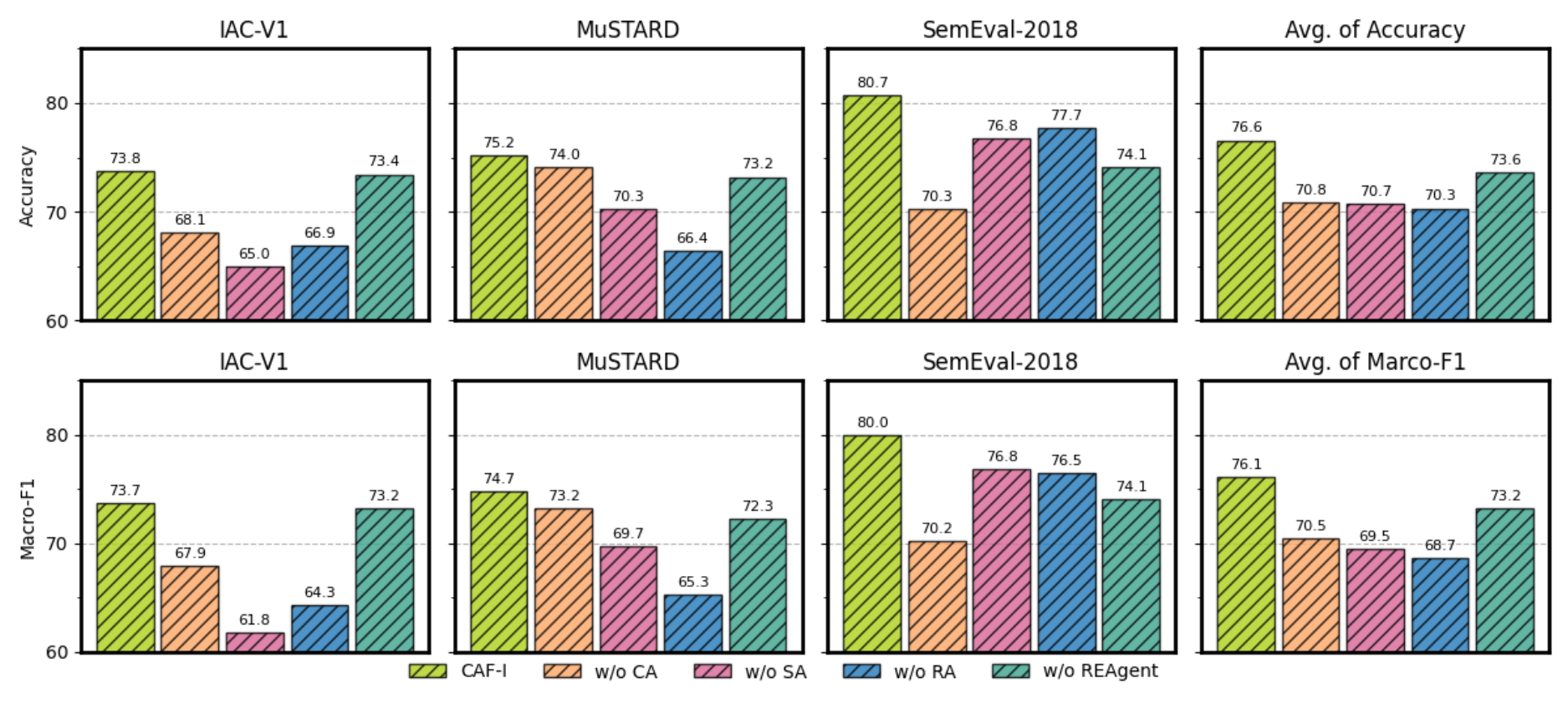}
    \caption{The ablation experimnet result of CA, SA, RA and REAgent.}
    \label{ablation}
    \vspace{-.6cm}
\end{figure}

To quantitatively assess the contribution of each core component within CAF-I, we conducted ablation studies, with detailed results presented in Figure~\ref{ablation}. These studies involved systematically removing specialized analysis agents CA, SA, or RA from all analysis stages, or deactivating the REAgent's conditional refinement mechanism, thereby making the initial aggregated decision final. Experiments were performed on the IAC-V1, MuSTARD, and SemEval-2018 datasets.

The results affirm the integral role of every component. Removing any single analysis agent consistently led to significant performance degradation across all datasets, underscoring their critical contributions. For instance, ablating the RA typically incurred a substantial average drop in Macro-F1. Similarly, deactivating the REAgent's refinement mechanism also noticeably reduced performance, particularly on more challenging datasets like SemEval-2018, confirming the value of this step for enhancing decision robustness. These findings validate the necessity of each specialized agent and the refinement process within CAF-I.

\begin{table*}[!htbp] 

    \centering
    \caption{Robustness evaluation of the CAF-I framework using different LLM backbones Qwen 2-7B, Llama 3-8B compared to the IO prompting. Baseline IO results are aligned with SarcasmCue \cite{yao2024sarcasm}. All scores are in \%.}
    \label{table:robustness_results}
    \scriptsize
    \setlength{\tabcolsep}{4pt}
    \begin{tabular}{l|cc|cc|cc|cc}
        \toprule
        \multirow{2}{*}{\textbf{LLM Backbone}} & \multicolumn{2}{c}{\textbf{IAC-V1}} & \multicolumn{2}{c}{\textbf{IAC-V2}} & \multicolumn{2}{c}{\textbf{SemEval-2018}} & \multicolumn{2}{c}{\textbf{MuSTARD}} \\
        \cmidrule(lr){2-3} \cmidrule(lr){4-5} \cmidrule(lr){6-7} \cmidrule(lr){8-9}
         & Acc. & Ma-F1 & Acc. & Ma-F1 & Acc. & Ma-F1 & Acc. & Ma-F1 \\
        \midrule
        \textbf{Qwen2-7b + IO} & 56.56 & 49.32 & 51.85 & 38.57 & 45.15 & 38.83 & 54.78 & 46.17 \\
        \textbf{Qwen2-7b + CAF-I} & \textbf{71.85} & \textbf{71.19} & \textbf{67.08} & \textbf{66.39} &\textbf{81.63} & \textbf{81.25}  & \textbf{72.60} & \textbf{73.80} \\ 
        \midrule
        \textbf{Llama3-8b + IO} & 55.94 & 46.40 & 54.70 & 43.74 & 49.36 & 44.46 & 54.64 & 44.99 \\ 
        \textbf{Llama3-8b + CAF-I} & \textbf{60.62} & \textbf{60.40} & \textbf{70.36} & \textbf{73.55} & \textbf{75.76} & \textbf{75.47} & \textbf{59.56} & \textbf{57.11}
        \\ 
        \bottomrule
    \end{tabular}%
    \vspace{-.6cm}
\end{table*}

\subsection{Robust Experiment}
To assess CAF-I's architectural robustness beyond its GPT-4o backbone, we tested it with Qwen 2-7B and Llama 3-8B on four benchmark datasets, maintaining the core framework and prompts. Performance was compared against each LLM's baseline IO prompt results, sourced from SarcasmCue \cite{yao2024sarcasm}, to isolate architectural contributions. Detailed results are in Table~\ref{table:robustness_results}.

The evaluation validates CAF-I's architectural value. Using alternative LLM backbones, CAF-I consistently yielded substantial average Macro-F1 improvements over their respective IO prompting baselines. Notably, on SemEval-2018, CAF-I with Qwen 2-7B achieved 81.25\% Ma-F1, even outperforming its primary GPT-4o backbone implementation which scored 79.99\% Ma-F1. This observation, alongside the general trend of outperforming IO baselines with different LLM, strongly confirms CAF-I's robustness and broad applicability.

\subsection{Explainability Experiment}
\begin{figure*}[!htp]
    \vspace{-.4cm}
    \centering
    \includegraphics[width=1.0\linewidth]{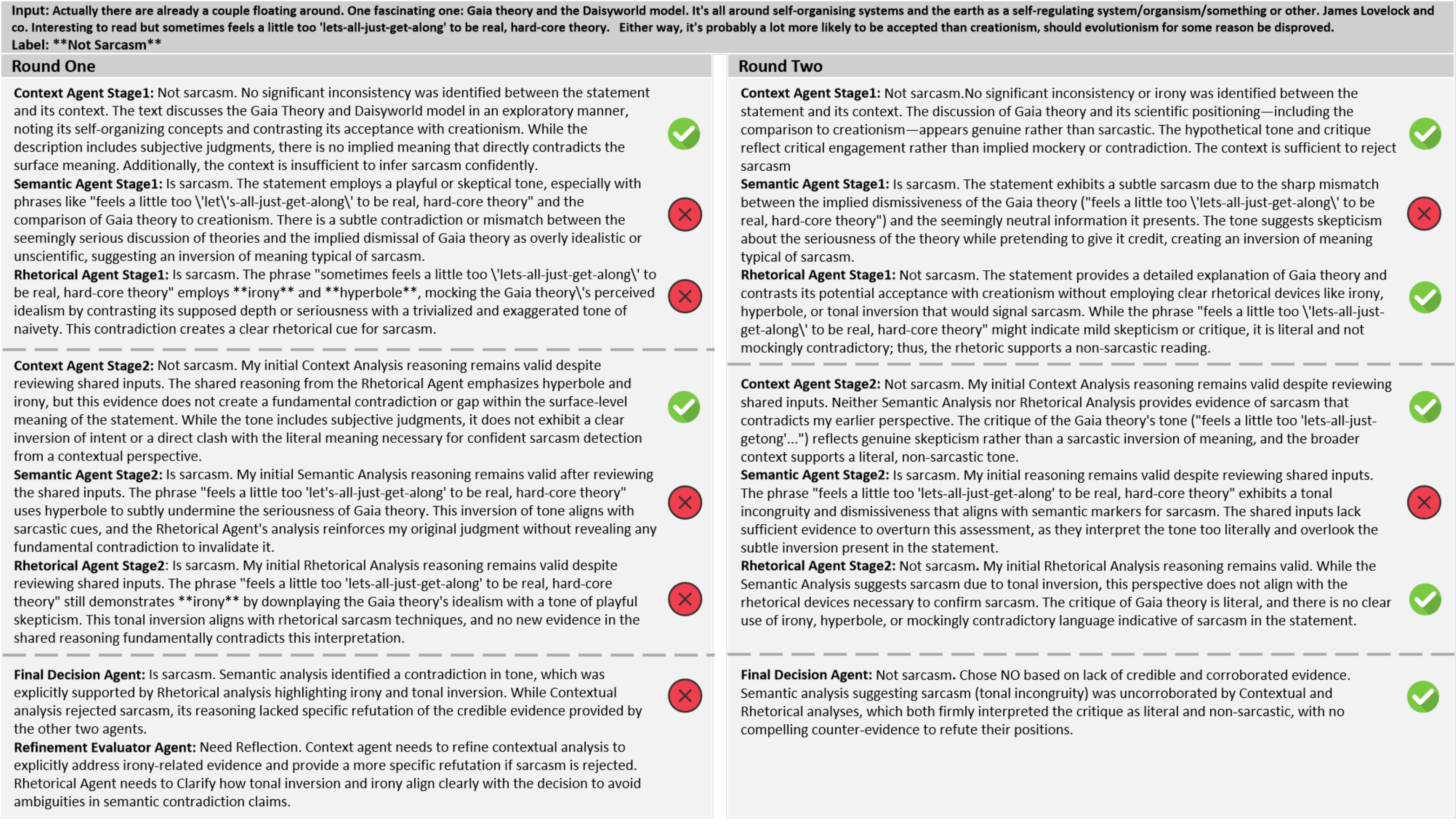}
    \caption{Illustration of the CAF-I framework's step-by-step reasoning and refinement process on a input example.}
    \label{fig:case_study_example}
        \vspace{-.6cm}

\end{figure*}
\subsubsection{Case Study}

Initially, an incorrect "Sarcasm" judgment arose because the Semantic and Rhetoric Agents over-interpreted certain phrases, despite the Context Agent finding no significant inconsistency. However, the REAgent identified this discrepancy and triggered a conditional refinement. In the subsequent feedback-guided re-evaluation, the Rhetoric Agents revised their assessments, aligning with the Context Agent towards "Not Sarcastic." This revised consensus allowed the Decision Agent to correctly classify the input, demonstrating REAgent's effectiveness in error correction and enhancing judgment robustness.

\subsubsection{Quantitative Evaluation of Explanations}

An auxiliary experiment on the SemEval-2018 dataset, using GPT-3.5 Turbo as a baseline, assessed the utility of intermediate reasoning from CAF-I's analysis agents. We compared GPT-3.5 Turbo's performance under two conditions: using only the original input text with a standard IO prompt, versus augmenting the input with a feature prompt concatenating stage 1 textual explanations from CAF-I's Context, Semantic, and Rhetoric agents. Incorporating these agent explanations improved the baseline model's \textbf{Macro-F1} score from \textbf{68.4\%} to \textbf{70.2\%}. This performance gain indicates that CAF-I's specialized agents generate valuable, discriminative reasoning, substantiating the framework's effective internal processing and enhancing its interpretability.

\subsection{Inference Efficiency Analysis}
\label{sec:inference_efficiency}
To assess practical trade-offs, we analyzed CAF-I's inference efficiency and detection performance against standard CoT and ToT baselines, all using the gpt-4o-mini backbone for fair comparison. This evaluation, detailed in Table~\ref{tab:inference_time_comparison}, used a 400-sample subset from our four benchmarks.

Table~\ref{tab:inference_time_comparison} indicates CAF-I achieves the highest accuracy and Macro-F1 scores. Its average inference time of 9.67 seconds per sample is highly competitive, nearly identical to that of ToT at 9.64 seconds and only marginally more than standard CoT at 8.93 seconds, despite CAF-I's sophisticated multi-agent architecture.

These results demonstrate CAF-I not only significantly outperforms both baselines in detection accuracy but does so without a substantial increase in computational overhead. This highlights the efficiency of CAF-I's collaborative reasoning design, leveraging architectural depth for superior accuracy while maintaining practical inference speed.

\begin{table}[h!]
    \vspace{-.2cm}

\centering
\caption{The average inference time (s/sample) and corresponding detection performance compared to baselines, using gpt-4o-mini on a 400-sample subset.}
\label{tab:inference_time_comparison}
\setlength{\tabcolsep}{6pt}
\renewcommand{\arraystretch}{0.8}
\begin{tabular}{l|c|c|c}
\toprule
Method & Acc.& Ma-F1 & Avg. Inf Time \\
\midrule
CoT  & 63.50 & 62.48 & 8.93\\
ToT    & 68.75 & 66.44 & 9.64 \\ \midrule
\textbf{CAF-I}   & \textbf{72.50} & \textbf{71.03} & 9.67 \\
\bottomrule
\end{tabular}
    \vspace{-.4cm}

\end{table}

\section{CONCLUSION}
This paper introduced \textbf{CAF-I}, an LLM-driven multi-agent framework for robust irony detection, integrating specialized context, semantic, and rhetoric agents with collaborative refinement. Experiments confirmed CAF-I's state-of-the-art zero-shot performance over existing methods. Ablation studies validated its component necessity and architectural robustness, while also highlighting its interpretability and error correction.

\bibliographystyle{splncs04}
\bibliography{mybibliography}

\end{document}